\documentclass[11pt]{article}

\usepackage[final]{acl}

\usepackage{times}
\usepackage{latexsym}

\usepackage[T1]{fontenc}

\usepackage[utf8]{inputenc}

\usepackage{microtype}

\usepackage{inconsolata}

\usepackage{graphicx}

\usepackage{booktabs}
\usepackage{multirow}
\usepackage{tcolorbox}

\title{Which one is banana man? Evaluating vision--language models in multi-turn pragmatic interpretation}

\author{
 \textbf{Alvin W. M. Tan*\textsuperscript{1}},\;
 \textbf{Ben Prystawski*\textsuperscript{1}},\;
 \textbf{Veronica Boyce\textsuperscript{1,2}}
\vspace{0.3em} \\ 
 \textsuperscript{1}Department of Psychology, Stanford University, USA\\
 \textsuperscript{2}Department of Brain and Cognitive Sciences, MIT, USA\\
 {*}Equal contribution
\vspace{0.3em} \\
\texttt{\{tanawm, benpry\}@stanford.edu\qquad vboyce@mit.edu}
}

\begin{document}
\maketitle
\begin{abstract}
Flexible adaptation to context and shared pragmatic intuitions contribute to smooth human conversation. 
Iterated reference games---in which players repeatedly pick out novel referents using language---present a test case for agents' ability to perform context-sensitive pragmatic reasoning in multi-turn linguistic environments. 
We tested humans and vision--language models on their ability to identify the intended meaning of descriptions produced in iterated reference games, varying the provided context in terms of amount, order, and relevance. 
While humans performed well consistently, the models we evaluated could make use of prior context to interpret humans' referring expressions, but they struggled to build up the relevant context to interpret those expressions effectively. 
Our results suggest that the models we evaluated lack core skills needed for efficient linguistic collaboration.
\end{abstract}

\section{Introduction}

Recent advances in machine learning have produced multi-turn conversational agents, which are a public face of artificial intelligence \cite{openaiIntroducingChatGPT2024}.
The utility of conversational agents depends on capacities including natural language understanding, world knowledge, and instruction following \cite{guanEvaluatingLLMbasedAgents2025}.
Furthermore, success in multi-turn conversation requires the underlying language model to respond to the user's message appropriately given the preceding context, which in turn requires the model to retain relevant contextual information and use it to interpret new messages. 

Such multi-turn interactions are a core feature of human communication: shared conversational history supports a shared system of semantic meaning \cite{clarkUsingLanguage1996, geurtsCommonGroundPragmatics2024}. 
\textit{Iterated reference games} are a common experimental paradigm for studying this type of communication. 
In iterated reference games, a describer produces a description of a referent so that a matcher can select the referent from a set of options. 
Over multiple rounds, conventionalised referring expressions usually emerge from repeated descriptions of the same referent \cite{clarkReferringCollaborativeProcess1986}.
These experiments demonstrate that humans dynamically adapt to their conversational partners, creating ad hoc context- and/or partner-specific meanings and conventions \cite{clarkReferringCollaborativeProcess1986, hawkinsPartnersPopulationsHierarchical2021, boyceInteractionStructureConstrains2024}. Humans can also comprehend conventionalized descriptions from games they were not part of, especially when provided with the original context
\cite{schoberUnderstandingAddresseesOverhearers1989, hawkinsVisualResemblanceInteraction2023, boyceIdiosyncraticNotOpaque2025}.

Conversations---whether human--human or human--AI---require their participants to use contextual information to understand the meanings of utterances.
A system that uses context to understand conversation would show two behavioural signatures: (1) the ability to interpret linguistic meaning pragmatically and (2) sensitivity to contextual information \cite{hawkinsVisualResemblanceInteraction2023, gulCoGenLearningFeedback2024, huaTalkLessInteract2024, sravanthiPUBPragmaticsUnderstanding2024, boyceIdiosyncraticNotOpaque2025, chenRetrospectiveLearningInteractions2025, junkerSceneGramConceptualizingDescribing2025}. 

In this work, we employ iterated reference games as a minimal test case to investigate whether state-of-the-art vision--language models (VLMs) reason pragmatically about referring expressions to resolve referential ambiguity given varying types of prior context, and how their sensitivity to context compares to that of humans. 

\section{Related work}

Model performance in interpreting descriptions from iterated reference games relies on several capacities, including the use of visual information for grounding, tracking and deploying information over multi-turn conversation, pragmatic reasoning, and conversational adaptation.

\subsection{Integrating visual information in VLMs}
Reference games use images to provide grounded context, but the vision abilities of VLMs are much weaker than their language understanding abilities. 
Some image information encoded by the vision encoder is not effectively decoded and used by VLMs, leading to poor performance at image classification for lower frequency classes \cite{zhang2024}, poor performance at identifying shape overlaps or counting shapes \cite{rahmanzadehgerviVisionLanguageModels2025}, and poor spatial reasoning \cite{kamath2023, liu2023a, chenWhySpatialReasoning2025}. 
VLMs also tend to overlook atypical features of images and rely on general world knowledge instead \cite{voVisionLanguageModels2025}. 
Thus, VLMs seem to remain limited in their ability to flexibly integrate relevant visual information \cite{tongLanguageModelingExploration2026}. 

\subsection{Evaluating multi-turn conversational abilities in language models}

Several recent benchmarks, such as MT-Eval \cite{kwan2024}, MultiChallenge \cite{sirdeshmukhMultichallengeRealisticMultiturn2025}, Multi-Turn Puzzles \cite{badola2025}, and MT-PingEval \cite{eisenstein2026mt}, have evaluated language models in multi-turn contexts to assess their ability to maintain coherent responses across a conversation and use earlier-provided information to inform later responses. 
Even contemporary commercial models do poorly on these benchmarks, exhibiting incoherence across responses, and failure to use earlier information when it is relevant but not explicitly noted as such. 
Models also struggle to efficiently collaborate to establish common ground \cite{eisenstein2026mt}.
Model behaviour in multi-turn contexts is weaker than in single-shot contexts in part because errors can compound across the interaction; models perform better given correct context than their own previously generated responses \cite{kwan2024}.

\subsection{Assessing pragmatics in VLMs}
Pragmatics is a broad term that encompasses language that is sensitive to different types of context, including cultural knowledge, grounded visual context, the knowledge state of the interlocutor, and the recent conversational context \cite{floyd2025}. 
Grounded referring expressions are a long-standing test bed for building computational models of pragmatics, which are historically rule-based symbolic or probabilistic models \citep{golland2010game, krahmer2012computational, kazemzadeh2014referitgame}. Some of these models learned from co-occurrence statistics to overcome misaligned or incomplete perceptual information \citep{liu2012towards, tellex2014learning, liu2015learning}. More recently, single-shot reference games have been used to assess how models understand and generate contextually appropriate descriptions, and to evaluate whether they match human preferences, among other pragmatic tests \cite{fried2023, ma2025}.
Referring tasks can also test interlocutor-specific adjustments when a specific listener has a different visual perspective or a different set of background knowledge and vocabulary \cite{bao2022, takmaz2023, tang2024}. 

\subsection{VLMs and iterated reference games}
Iterated reference games represent a difficult test of VLM pragmatics, sometimes incorporating difficult visual context and requiring online adaptation to the conversational partner. 
Prior work suggests VLMs can do a reasonable job at comprehending human-generated descriptions of naturalistic images that increase in efficiency \cite{hawkinsContinualAdaptationEfficient2020, greco2023, huaTalkLessInteract2024}. 

Several recent studies have compared human--human dyads to human--AI and AI--AI dyads using more difficult contexts such as tangrams or similar-looking visual stimuli \cite{jones2026, zeng2026}. While frontier models can achieve reasonable accuracy when playing with each other, their dynamics and descriptions do not match the patterns of humans \cite{imai2025measuring, zhaoComparingHumanMachine}. Human--AI dyads perform poorly, with low accuracy, compared to either AI--AI or human-human dyads. 

Training models on the task and giving explicit prompts both lead to somewhat higher performance and superficially more human-like descriptions than base models \cite{takmaz2020, gulCoGenLearningFeedback2024, huaTalkLessInteract2024, chenRetrospectiveLearningInteractions2025, vaduguru2025a}, but no model shows the partner-sensitive convention-formation characteristic of human interaction. 

While they did not compare directly to human performance, \citet{wangLVLMsAreBad2025} find that VLMs are generally weak at understanding descriptions for iterated reference games, regardless of the amount of earlier context. 

\begin{figure*}[th]
    \centering
    \includegraphics[width=\textwidth]{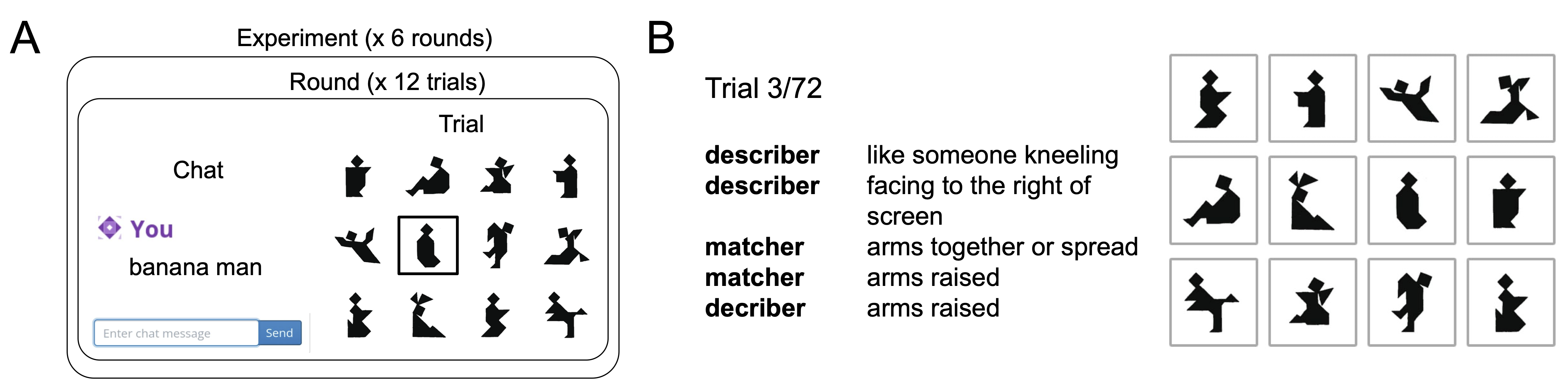}
    \caption{(A) Overview of the experimental structure for the original interactive games in \citet{boyceInteractionStructureConstrains2024}. (B) User interface for each trial in the experiments with na\"ive participants.}
    \label{fig:setup}
\end{figure*}

\section{Methods}

All model and analysis code is available on \href{https://github.com/benpry/vlm-tg-context-v2}{GitHub}, and data are available on \href{https://osf.io/zk8gq/}{OSF}.
AI assistants were used for some coding tasks, but not for any other purpose in this project.

\subsection{Dataset}

We used an iterated reference game dataset from \citet{boyceInteractionStructureConstrains2024}. 
In each game, players saw a grid of 12 tangram images (Figure \ref{fig:setup}A).
On each trial, the describer saw one image highlighted and described it to the other players via chatbox so that matchers could each select the target from the 12 options. 
Each game had 2--6 players, who played 6 rounds of 12 trials each (one trial for each image). 

Here we use ten of these games, which were subsequently used by \citet{boyceIdiosyncraticNotOpaque2025} to test for context sensitivity in naïve human matchers. 
Na\"ive matchers read transcripts of the conversations between the players in the original games and chose what image was being described (Figure \ref{fig:setup}B). 
They received feedback on whether they were correct, but were not told the correct answer if they were incorrect. 
\citet{boyceIdiosyncraticNotOpaque2025} collected participants who saw the trials in the original (yoked) order ($N$ = 99) or shuffled ($N$ = 98). 
We collected additional human data on Prolific in the backward ($N$ = 89) and random ($N$ = 107) conditions (described in Section~\ref{sec:contexts}) using the same procedure.
See Appendix~\ref{app:human} for more details.

\subsection{Experiment setup}

Our experiments with VLMs aimed to match the na\"ive human setup.
We focused on evaluating the instruction-tuned versions of five leading open-weights VLMs of different sizes: Qwen 3 VL 32B \cite{bai2025qwen3vltechnicalreport}, Gemma 3 27B \cite{team2025gemma}, Llama 3.2 11B \cite{llamateamLlama32Revolutionizing2024}, Molmo 2 8B  \cite{clark2026molmo2openweights}, and Kimi VL A3B \cite{kimiteamKimiVLTechnicalReport2025}.
Models were given a general system prompt describing the task setup, followed by the set of 12 tangram image options labeled A--L presented as a single image. All open models were run with the vLLM inference framework \citep{kwon2023efficient}. See Appendix~\ref{app:add_methods} for additional details.
Models were then given the context of all preceding trials formatted as a chat history, with the user supplying the text written by participants and the model serving as the assistant providing the target tangram, denoted A--L.
Finally, the description for the test trial was provided, and the model log probabilities for the tokens A--L were obtained.
We renormalised the log probabilities using a softmax, and treated the probability assigned to the correct target as the model's accuracy.

\begin{table*}[th]
    \small
    \centering
        \caption{Conditions of in-context trials seen by models. Columns indicate whether the in-context trials are from the same original game as the test trial, whether the in-context trials are all from the same original game, what the order of the in-context trials is, and whether the tangram in the test trial occurs in the in-context trials.}
        \label{tab:conditions}
    \begin{tabular}{llccccc}
\toprule
&Condition    & $N_\textup{trials}$ & Original game             & Same game                 & Trial order                                             & Same tangram seen         \\ \midrule
\multirow{4}{7em}{Cumulative} & Yoked        & 720                 & \cellcolor[HTML]{94F186}Y & \cellcolor[HTML]{94F186}Y & \cellcolor[HTML]{94F186}Original                        & \cellcolor[HTML]{94F186}Y \\
& Backward     & 720                 & \cellcolor[HTML]{94F186}Y & \cellcolor[HTML]{94F186}Y & \cellcolor[HTML]{F18D86}Reversed                        & \cellcolor[HTML]{94F186}Y \\
& Shuffled     & 7056                & \cellcolor[HTML]{94F186}Y & \cellcolor[HTML]{94F186}Y & \cellcolor[HTML]{F18D86}Permuted                        & \cellcolor[HTML]{94F186}Y \\
& Random       & 7704                & \cellcolor[HTML]{F18D86}N & \cellcolor[HTML]{F18D86}N & \cellcolor[HTML]{F18D86}Permuted & \cellcolor[HTML]{94F186}Y \\ \midrule
\multirow{4}{7em}{Non-cumulative} &Ablated      & 720                 & \cellcolor[HTML]{94F186}Y & \cellcolor[HTML]{94F186}Y & \cellcolor[HTML]{94F186}Original & \cellcolor[HTML]{F18D86}N \\
& Other-within & 6480                & \cellcolor[HTML]{F18D86}N & \cellcolor[HTML]{94F186}Y & \cellcolor[HTML]{94F186}Original & \cellcolor[HTML]{94F186}Y \\
& Other-across & 6480                & \cellcolor[HTML]{F18D86}N & \cellcolor[HTML]{F18D86}N & \cellcolor[HTML]{94F186}Original & \cellcolor[HTML]{94F186}Y \\
& No context   & 720                 & \cellcolor[HTML]{F18D86}N & \cellcolor[HTML]{F18D86}N & \cellcolor[HTML]{F18D86}None                            & \cellcolor[HTML]{F18D86}N \\ \bottomrule
    \end{tabular}
\end{table*}

\subsection{Context conditions\label{sec:contexts}}


To investigate the role of prior contextual information, we considered eight conditions varying the amount and type of in-context trials (Table~\ref{tab:conditions}). 
The conditions varied in their fidelity to the context experienced by human participants in the original iterated reference games.

Our main set of conditions contained trials that were cumulative (i.e., in-context trials had been seen exactly in order), for which we had comparative na\"ive human data.
Three conditions contained trials drawn from a single original game: the \textit{yoked}, \textit{backward}, and \textit{shuffled} conditions presented all the trials in the original game, in the original, reversed, or permuted order respectively.
We also included a \textit{random} condition that fully shuffled across games and trial orders. 

To further explore how prior context informed VLM interpretation, we considered additional conditions for which the trials were not cumulative.
The \textit{ablated} condition presented trials in the original order, but in-context trials containing the same target as the test trial were removed.
Two other conditions involved in-context trials sampled from different original games than the test trial.
The \textit{other-within} condition drew all in-context trials from a single different original game than the test trial, whereas the \textit{other-across} condition drew in-context trials randomly from all other original games.
Finally, we had a baseline condition of \textit{no context}, in which no in-context trials were presented.

\subsection{Feedback availability}
How much do models rely on information provided in feedback? We varied the type of feedback provided in the in-context trials to distinguish retrieval from learning.\footnote{Note that the no context condition did not include any in-context trials, so there was no effect of feedback condition.}
Our primary type of feedback was \textit{interactive limited feedback}: in-context trials contained models' own prior guesses, and whether those guesses were correct or not. 
This setup was the most similar to what na\"ive humans experienced in our experiments.
As additional comparisons, we also evaluated models given \textit{human-yoked limited feedback}: in-context trials contained human participants' prior guesses, and whether those guesses were correct or not; and given \textit{full feedback}: in-context trials contained the correct answers, always marked as correct.
In full feedback, many trials can be solved by retrieving the most semantically similar in-context trial and repeating the answer (whereas this heuristic may not have been available in the other feedback types since the previous guess may not have been correct).

\section{Results}

We first compare model and human performance across different trial orders and different amounts of prior context. We then use additional control conditions to more precisely evaluate models. 

\begin{figure*}[th]
    \centering
    \includegraphics[width=\linewidth]{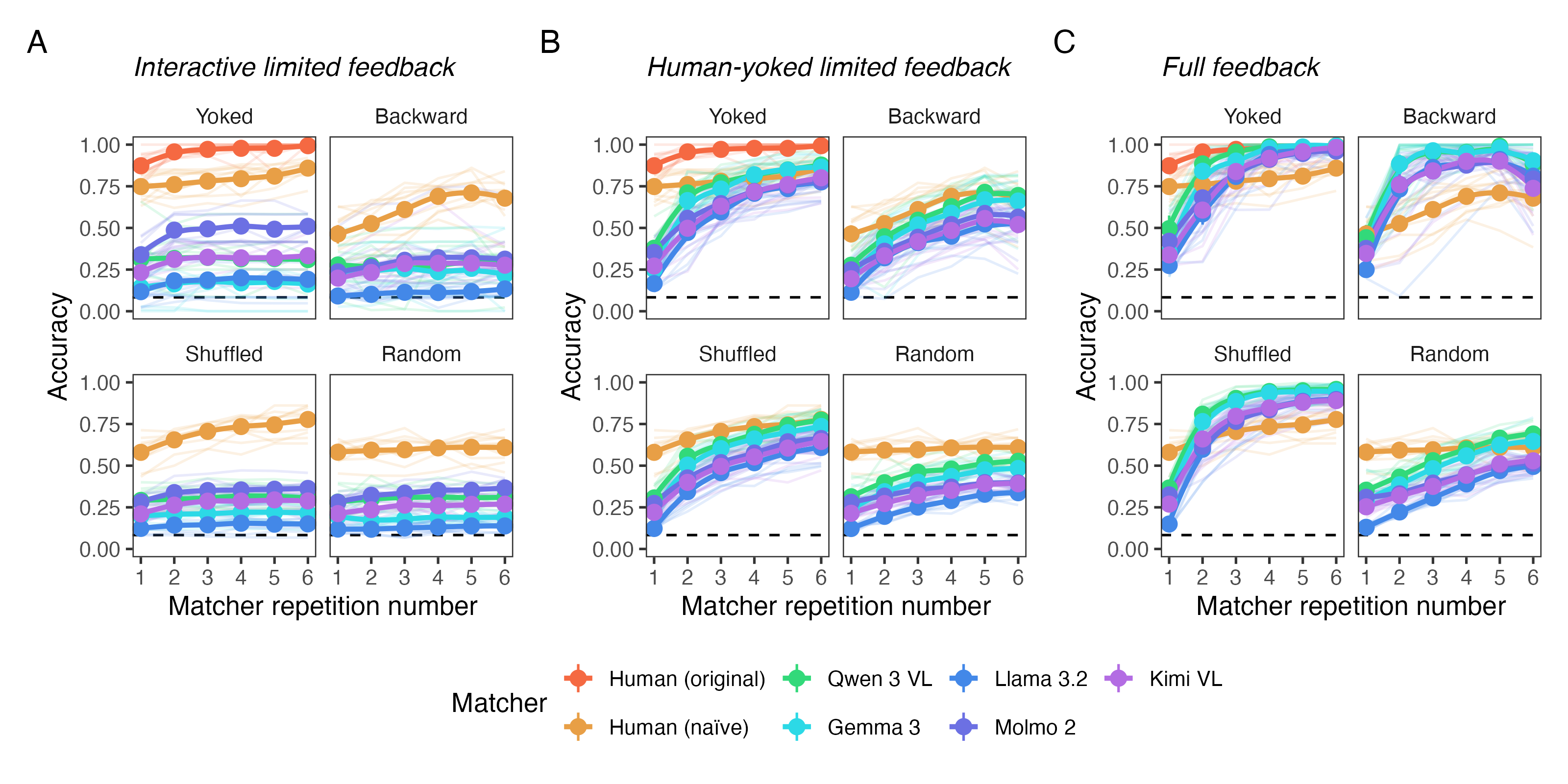}
    \caption{Matcher accuracy across all cumulative conditions and matcher types (both human and model), with best-fit LOESS curves, shown by repetition number as seen by the matcher. Subplots reflect the type of feedback received by model matchers: (A) interactive limited feedback, (B) human-yoked limited feedback, or (C) full feedback. Error bars indicate bootstrapped 95\% confidence intervals. Dashed lines indicate the chance level (0.083). }
    \label{fig:feedback}
\end{figure*}
\subsection{Models struggle to reason correctly in iterated reference games}
Our first question was whether VLMs were able to perform ad hoc pragmatic reference resolution in our abstract, out-of-domain context. 
Figure~\ref{fig:feedback}A shows the results from all open-weights models in the cumulative conditions under interactive limited feedback, along with comparative human data from the original reference games and from na\"ive humans.
All models demonstrated above-chance accuracy on the task across all conditions. 
Additionally, model performance began low and improved very slightly across rounds, especially from rounds 1 to 3, suggesting some adaptation to the task. 
In contrast, na\"ive humans' performance was relatively high even in the first repetition and showed greater improvement over rounds in the yoked, backward, and shuffled conditions. 
Performance in the shuffled and random conditions was generally poorer than in the yoked condition, matching na\"ive humans.
Broadly, interpreting pragmatic references to tangram images remains a challenging task for models in comparison to humans.

\subsection{Providing additional contextual information improves model performance}
We next examined how feedback availability affected model performance.
Figure~\ref{fig:feedback}B and \ref{fig:feedback}C show the results from all models in the cumulative conditions under human-yoked limited feedback and full feedback respectively, again with comparative human data (although it is important to note that the humans themselves only received interactive limited feedback). 
Models showed substantial improvement with human-yoked limited feedback, and further improvements still with full feedback.
Notably, in the yoked, backward, and shuffled conditions, models reached an asymptote close to human levels in the human-yoked limited feedback condition, and close to 100\% accuracy in the full feedback condition.
In the random condition, models mostly remained worse than humans; drawing in-context trials from different games may have limited the amount of relevant contextual information available.
These trends suggest that models were able to exploit the correct responses found in the in-context trials to attain better subsequent performance.

\subsection{Only relevant contextual information improves model performance}
\begin{figure}[t]
    \centering
    \includegraphics[width=\linewidth]{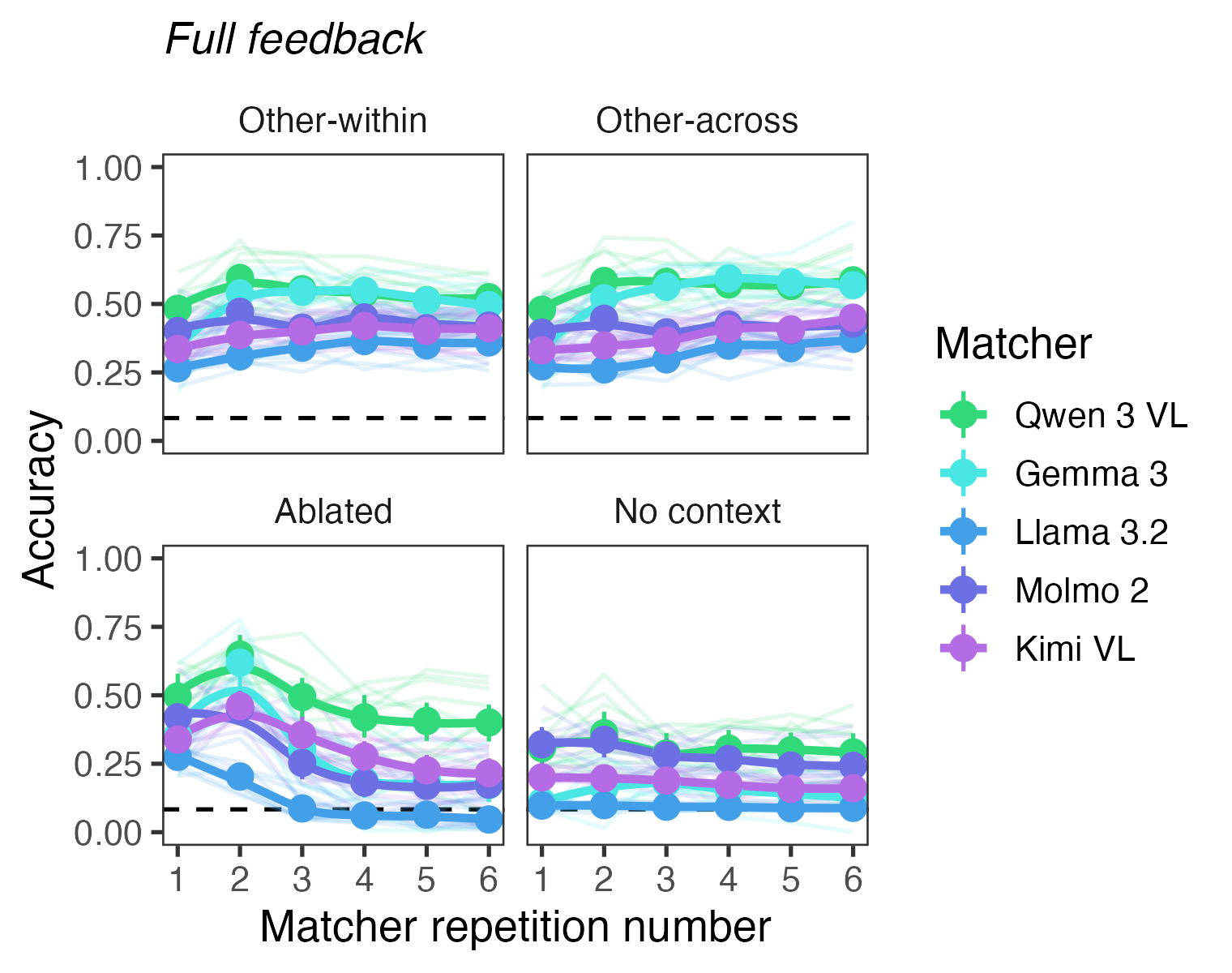}
    \caption{Matcher accuracy in full feedback across all non-cumulative conditions, with best-fit LOESS curves, shown by repetition number as seen by the matcher except for the no context condition, which is shown by the original repetition number. Error bars indicate bootstrapped 95\% confidence intervals. Dashed lines indicate the chance level (0.083).}
    \label{fig:noncumulative}
\end{figure}
\begin{figure*}[th]
    \centering
    \includegraphics[width=0.9\linewidth]{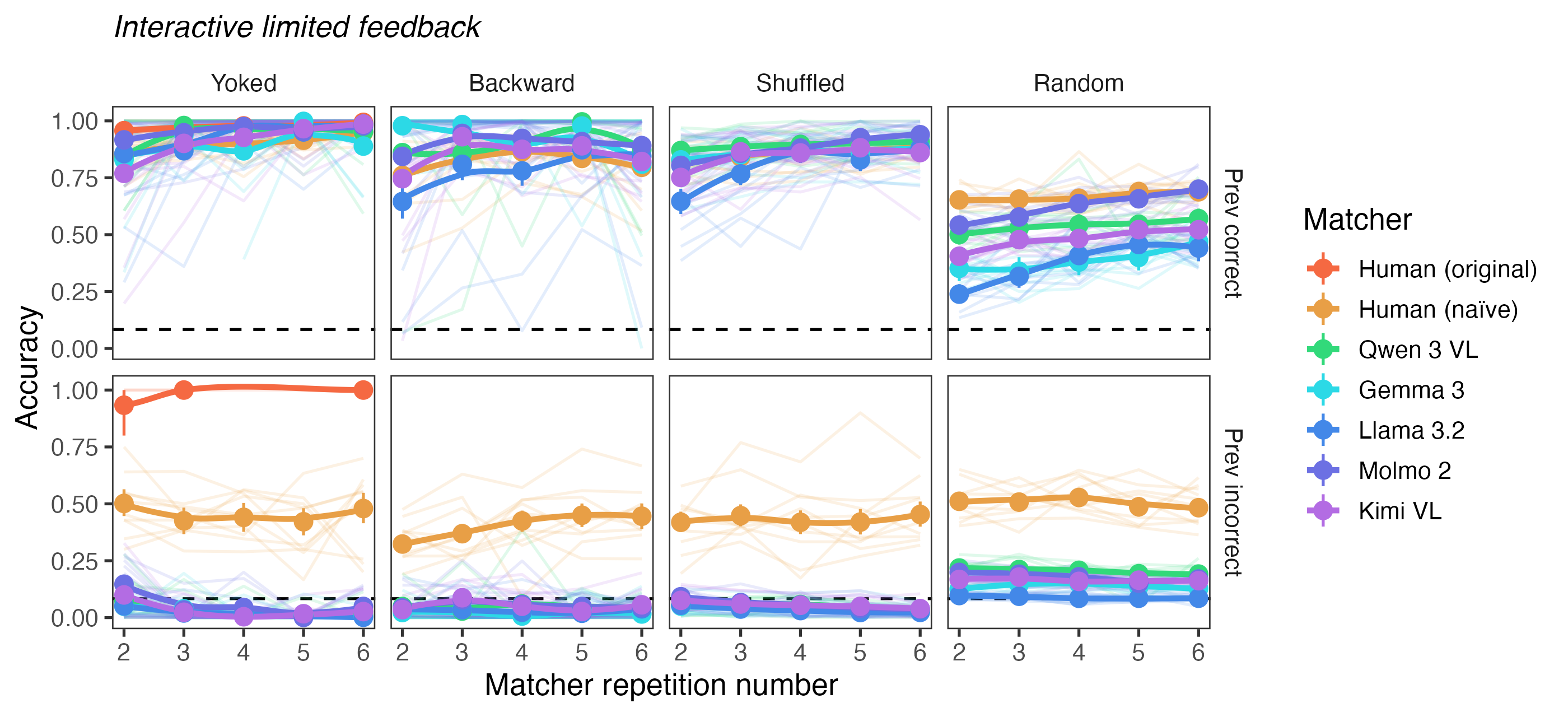}
    \caption{Trial accuracy as a function of previous-repetition accuracy for the same tangram across matchers for the yoked, backward, shuffled, and random conditions under interactive limited feedback. The top row displays accuracy for trials subsequent to a correct trial, while the bottom row displays accuracy for trials subsequent to an incorrect trial. Error bars indicate bootstrapped 95\% confidence intervals. Dashed lines indicate the chance level (0.083).}
    \label{fig:learning}
\end{figure*}
We investigated what types of contextual information help model performance by turning to the non-cumulative conditions under full feedback, as shown in Figure~\ref{fig:noncumulative}.
In these conditions, information relevant to the current trial was unlikely to be present in the in-context trials, since previous instances of the specific target--game combination of the test trial were not present. 
When the conversational history came from a different game or games (other-within and other-across), models had much lower accuracy of 0.3 -- 0.6. 
The lower performance compared to the yoked, shuffled, and backward conditions suggests that model improvement with relevant context was not simply adaptation to the task. 
Rather, the boost depended on the in-context and test trials coming from the same original game, as conventions reached by one game are not necessarily predicted by the context from other games. 
Performance was also low in the ablated condition, implying that experience with the test trial tangram itself was crucial to understand a convention, which was not systematically inferrable from other tangram conventions, even from the same game. 
These results further support the hypothesis that models are mostly retrieving from relevant in-context trials rather than learning over the course of the task.
Nonetheless, model performance in the other-within, other-across, and ablated conditions tended to be better than the no context condition, suggesting that at least some learning was occurring.

\subsection{Models fail to learn from past mistakes}\label{sec:learning}
As an alternative window into the effect of feedback on performance, we visualised matcher accuracy from the second repetition onward as a function of whether the matcher selected correctly in the previous repetition of the same tangram, as shown in Figure~\ref{fig:learning}.\footnote{We quantised previous repetition accuracy by determining whether the argmax of the probabilities was the target tangram for models, and by determining whether any matcher selected correctly for the original-game humans; na\"ive human performance was already binarised.}

In the yoked, shuffled, and backward conditions, when both humans and models answered a previous repetition correctly, they were also very likely to answer the next repetition correctly.
However, results were different when matchers answered a previous repetition incorrectly: na\"ive humans were able to correct their previous mistakes approximately half the time, whereas models were at or worse than chance.
Thus, models' overall performance may have been largely driven by the correct answers they happened to attain and retain, rather than learning over the course of the task.
Follow-up analyses where models saw the same items repeatedly demonstrate that models did not learn over repetitions, and instead perseverated, selecting the same wrong answer (see Appendix~\ref{app:practice}).
We speculate that the incorrect responses may have provided additional evidence for incorrect sequences, increasing the probability that they are later reproduced (ignoring the ``correct/incorrect'' feedback). 
Further work exploring perseveration across a broader range of contexts will help to clarify model response dynamics across multi-turn conversations.

\subsection{VLMs make limited use of visual information}
\begin{figure*}[th]
    \centering
    \includegraphics[width=0.9\linewidth]{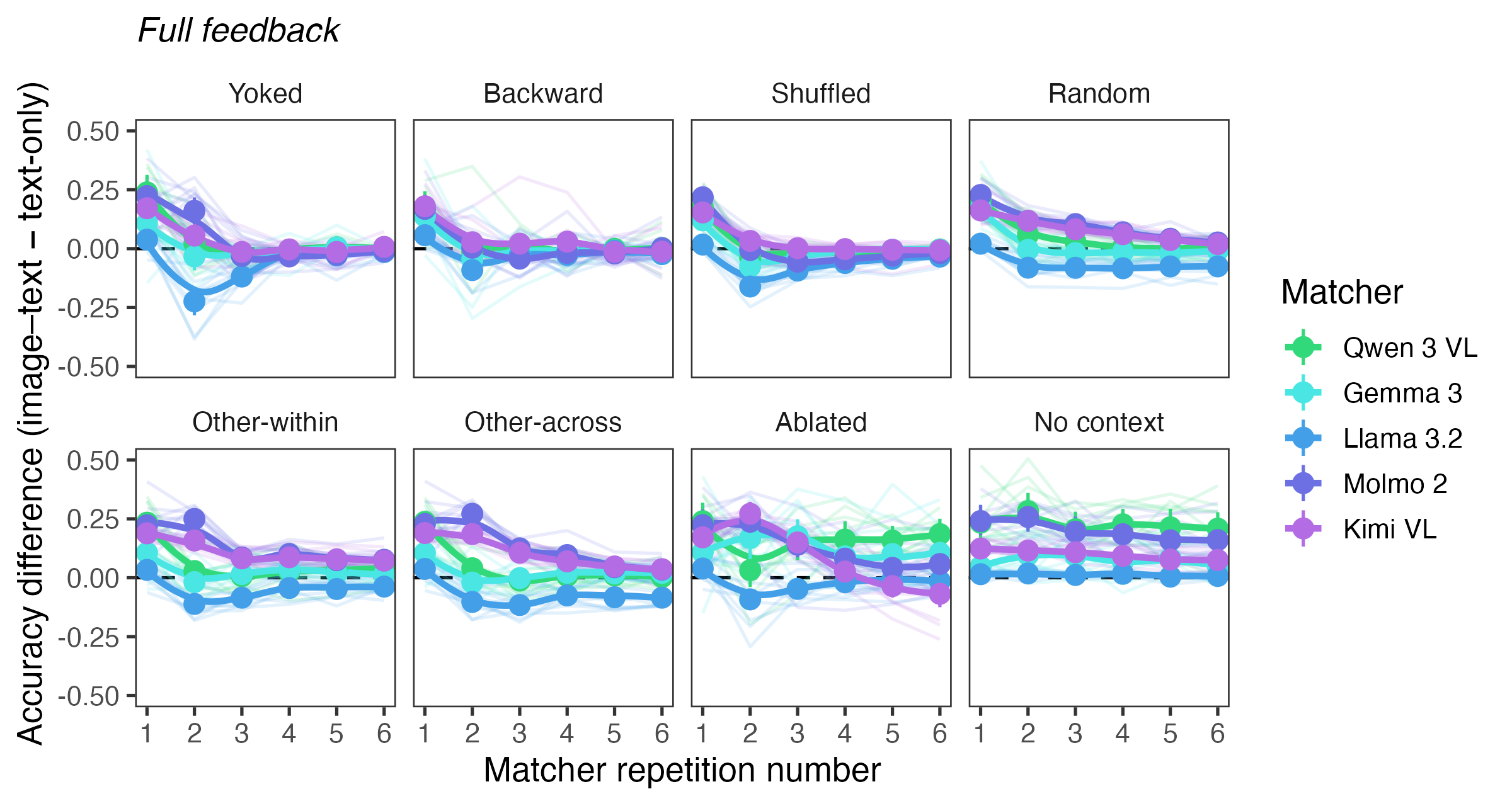}
    \caption{Difference in accuracy between the text-only and image--text paradigms across all conditions and models under full feedback. Error bars indicate bootstrapped 95\% confidence intervals. Dashed lines indicate no difference.}
    \label{fig:noimg_diff}
\end{figure*}
To determine the extent to which VLMs were using the tangram images to perform pragmatic reasoning, we also ran all models using a prompt that did not include the tangram images.
The difference in accuracy between the image--text and the text-only paradigms is shown in Figure~\ref{fig:noimg_diff}.

Models generally performed worse without images than with images in the first repetition, and performance collapsed to chance without context; however, by the third repetition, models were not different whether or not they had images in most conditions.
Models remained slightly worse without images in conditions with limited relevant contextual information (random, other-within, other-across, and ablated).
Notably Llama 3.2 was better without the image in many conditions, especially in the second repetition.
Some models seem to primarily use text information to select the correct match, which corroborates results from \citet{huaTalkLessInteract2024}, showing that when images are shuffled from trial to trial, VLMs exhibit little to no learning over repetitions.

\subsection{Models and humans diverge in their responses}\label{sec:spans}
\begin{figure*}[th]
    \centering
    \includegraphics[width=.9\linewidth]{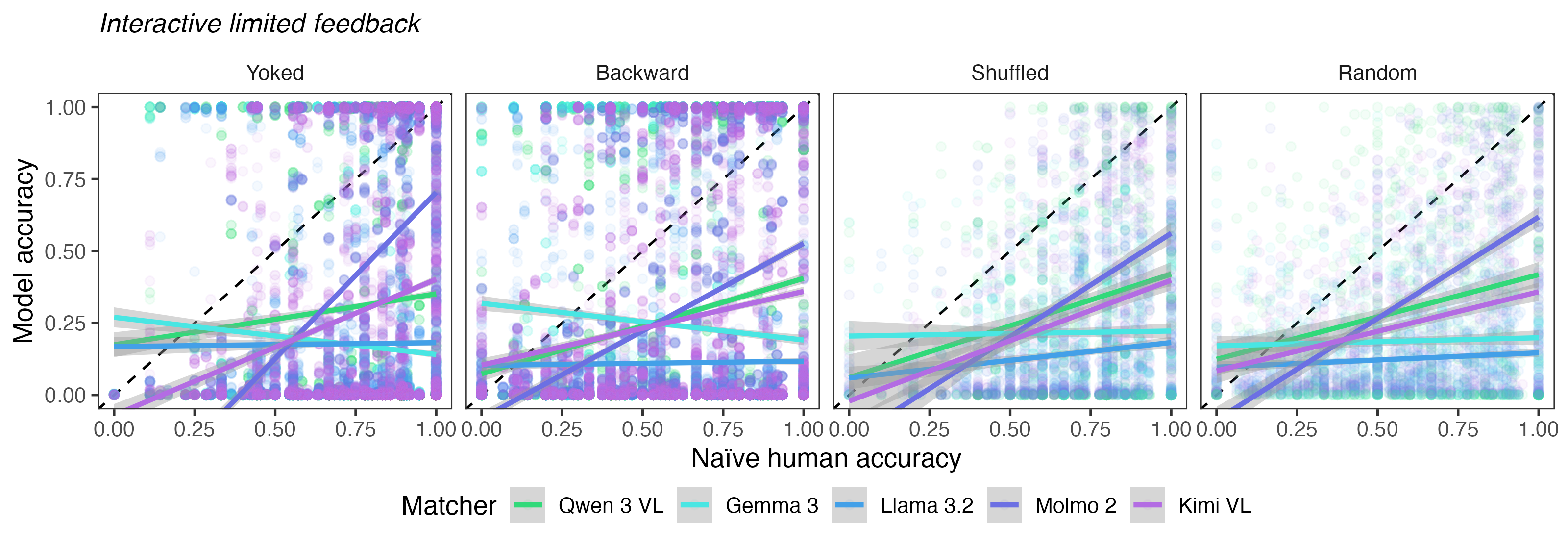}
    \caption{Comparison between naïve human accuracy and model accuracy under interactive limited feedback, with best-fit linear regressions. Shaded regions indicate 95\% confidence intervals. Dashed lines indicate perfect calibration ($y = x$).}
    \label{fig:comparison}
\end{figure*}

To examine whether models generally showed the same response patterns as humans, we compared model and na\"ive human performance at the trial level under interactive limited feedback (Figure~\ref{fig:comparison}).  
Models were relatively poorly calibrated, with weak correlations between model and human performance (yoked model $r$ = -.07 -- .46, human split-half $r$ = .42 [.32, .50]; backward model $r$ = -.07 -- .40, human split-half $r$ = .48 [.40, .55]; see Appendix~\ref{app:corrs} for more details).
Models may not have the same factors affecting their performance as humans, resulting in a lack of trial-wise similarity. 

We investigated the divergent human and model results by exploring features of the tangram descriptions driving human and model performance.
Specifically, we investigated the possible roles of metaphorical and holistic descriptions.
These dimensions were inspired by \citet{clarkReferringCollaborativeProcess1986}, who predicted that in later repetitions, dyads should prefer metaphorical descriptions to literal ones and holistic descriptions to part-based ones, as metaphorical and holistic descriptions would require less communicative effort.\footnote{Note that \citet{clarkReferringCollaborativeProcess1986} referred to metaphorical descriptions as ``analogical''.}
\citet{clarkReferringCollaborativeProcess1986} also suggested that these two dimensions may be strongly correlated---metaphorical descriptions allow for different components to be jointly described as an overall silhouette, whereas more literal descriptions typically juxtapose descriptions of the individual components. 
Since the later repetitions were when models and humans diverged the most, we hypothesised that the types of descriptions may play a role in explaining such divergence.
(See Appendix~\ref{app:lemmas} for a complementary approach focusing on lemmas.)

\begin{figure}[t]
    \centering
    \includegraphics[width=\linewidth]{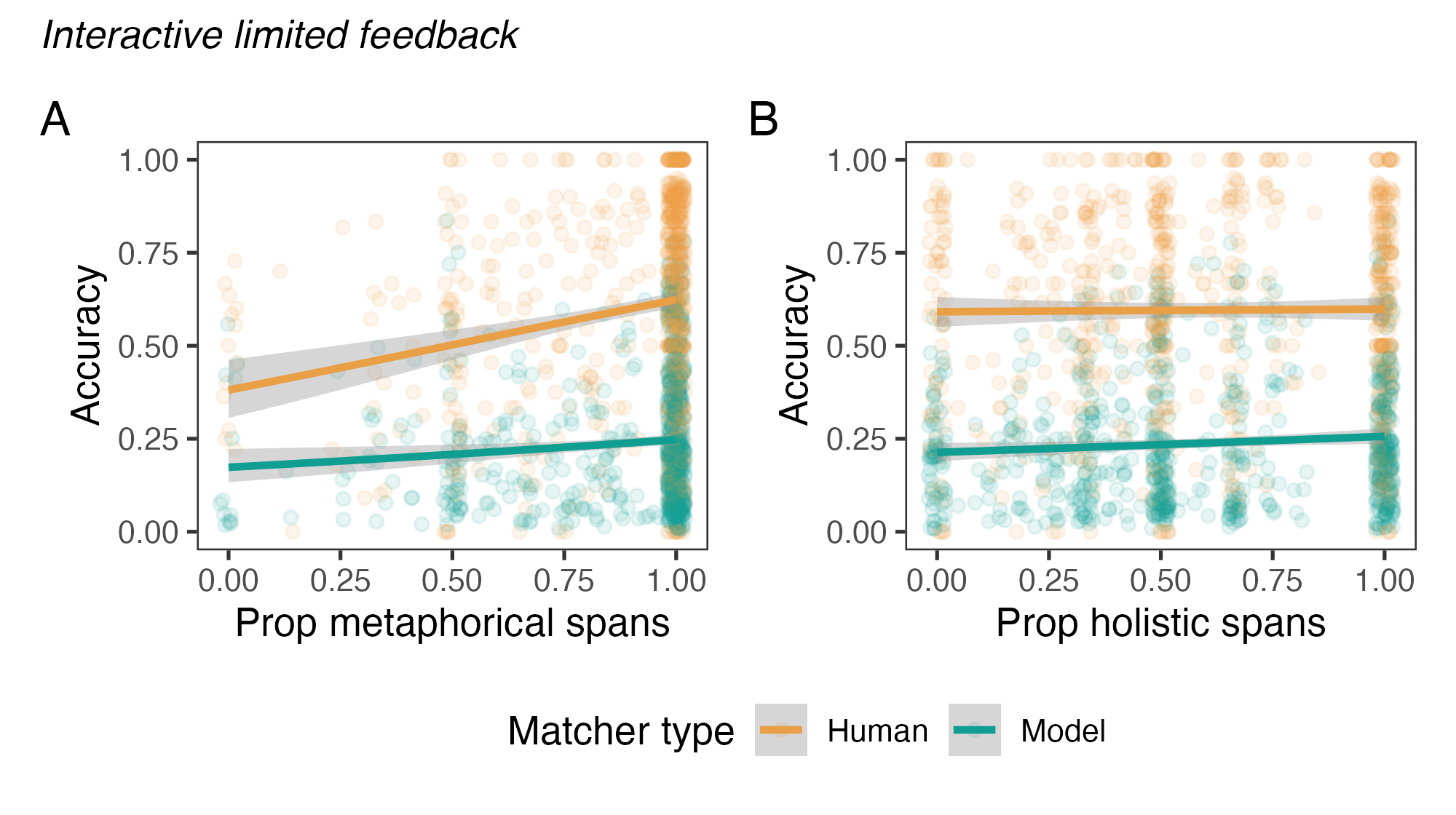}
    \caption{Matcher accuracy as a function of the proportion of metaphorical and holistic spans in descriptions.}
    \label{fig:spans}
\end{figure}

We used LLM-assisted annotation to annotate spans of referential descriptions in each tangram description; see Appendix~\ref{app:annotation} for more details.
The proportion of metaphorical spans and the proportion of holistic spans were significantly but only moderately correlated ($r$ = .36 [.32, .41], $p$ < .001), suggesting that they were indexing different features of descriptions.
We then ran a quasibinomial regression predicting matcher accuracy from matcher type (human / model, with human as the reference level), the proportion of metaphorical spans, the proportion of holistic spans, and all interactions among these predictors. 
As shown in Figure~\ref{fig:spans}, there was a significant effect of the proportion of metaphorical spans ($b$ = 1.38 [1.08, 1.69], $p$ < .001) as well as matcher type ($b$ = -0.90 [-1.20, -0.61], $p$ < .001).
Most interestingly, there was also a significant interaction between the proportion of metaphorical spans and matcher type ($b$ = -1.06 [\mbox{-}1.40, -0.72], $p$ < .001); estimated marginal trends showed that humans showed a strong positive effect of metaphorical spans ($b$ = 1.08 [0.84, 1.31]) whereas models showed a much weaker effect ($b$ = 0.22 [0.10, 0.35]).
No effects including the proportion of holistic spans were significant.

Additional analyses on item-level effects and condition-wise correlations are shown in Appendix~\ref{app:add_results}.

\begin{figure}[t]
    \centering
    \includegraphics[width=\linewidth]{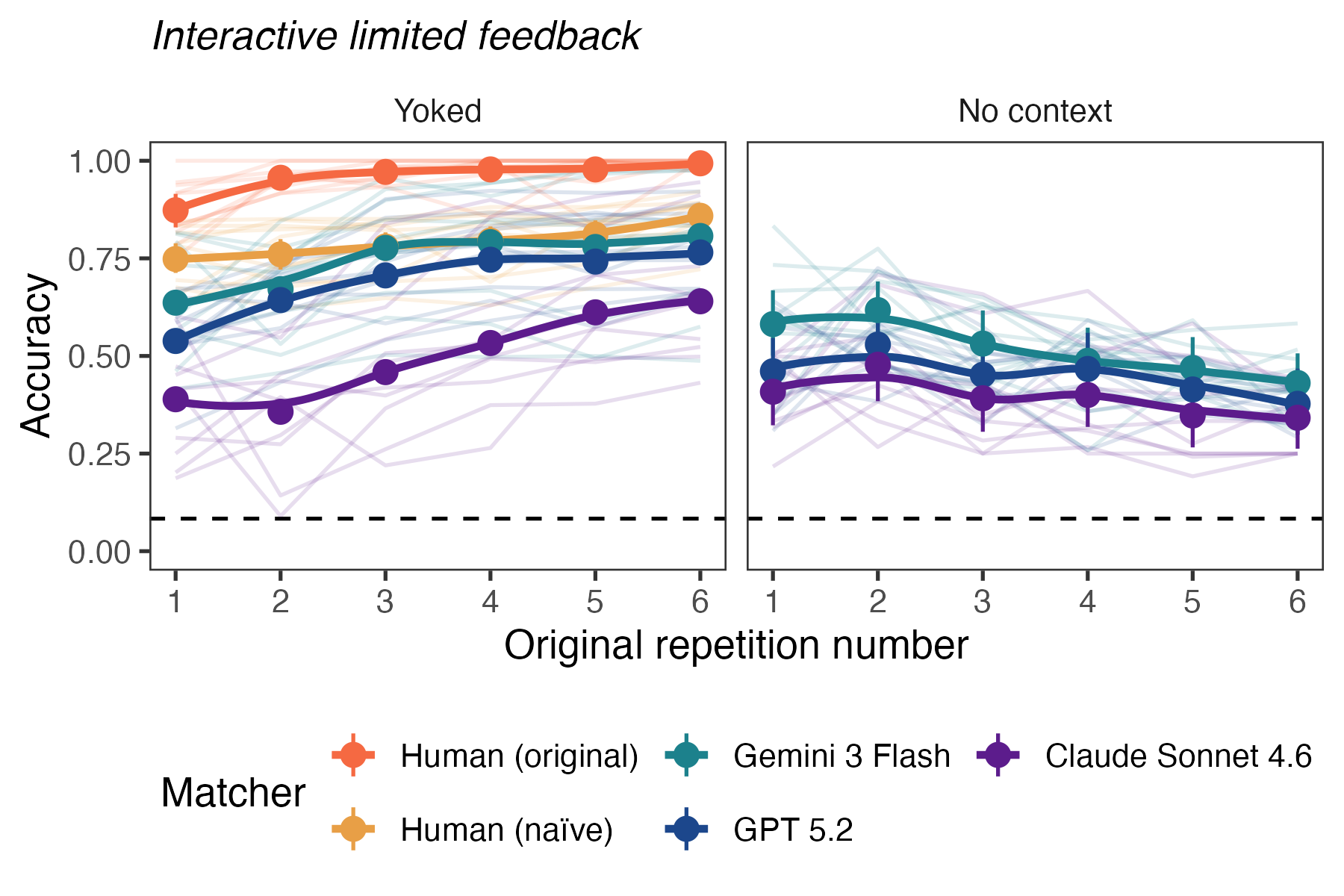}
    \caption{Matcher accuracy for frontier models under interactive limited feedback.}
    \label{fig:frontier}
\end{figure}

\subsection{Frontier models still fail to reach human-level performance}
As an extension, we evaluated three frontier VLMs on the yoked and no context conditions under interactive limited feedback to determine whether better and larger models trained on more data could approach human-level performance.
The frontier models we evaluated were Gemini 3 Flash \cite{geminiNewEraIntelligence2025}, GPT 5.2 \cite{openaiIntroducingGPT522025}, and Claude Sonnet 4.6 \cite{claudeteamIntroducingSonnet462026}.
The log probabilities were not fully obtainable for all of these models, so we computed a Monte Carlo estimate of their probability distributions by resampling generations 10$\times$ for each trial (see Appendix~\ref{app:inference} for more details).
We used the lowest reasoning level to approximate humans' relatively fast response times on this task.
Models sometimes failed to return valid responses; we ignored such errors and renormalised excluding them (see Appendix~\ref{sec:invalid}).

As shown in Figure~\ref{fig:frontier}, frontier models demonstrated better performance than open-weights models in general, but were still worse than na\"ive humans on the whole. 
The no context condition had only slightly worse accuracies than the yoked condition, also suggesting limited learning over the course of the task.
Trials from later rounds were also more difficult for frontier models.

\section{Discussion}

Multi-turn conversation is a key aspect of human communication and a challenge for AI agents.
We demonstrated that state-of-the-art open-weights VLMs were able to perform ad hoc pragmatic reasoning across multiple turns of a reference game, but were still substantially worse than humans. 
Models were sensitive to the presence of relevant contextual information, showing greater improvement when correct answers were present in their context, and limited learning when the correct answers were unavailable. 
This pattern appeared due to  models' failure to learn from past mistakes, whereas humans could.
Our results align with previous work on in-context learning, where increasing the number of valid in-context examples improved model performance, especially on difficult tasks \cite{chenHowManyDemonstrations2023, agarwalManyShotInContextLearning2024, jiangManyShotInContextLearning2024}.

Furthermore, models displayed different patterns of performance than humans and showed weak trial-wise calibration to human accuracies. 
These discrepancies were partially explained by the fact that humans were much more accurate with more metaphorical spans, whereas models were only weakly so.
The differential sensitivity to quantity and type of context between humans and models could suggest different approaches to the task. 
For example, models can (theoretically) exactly retrieve previous trial descriptions, whereas humans have more limited working memories and may need to use compression or other strategies to recall appropriate description--referent mappings. 

Overall, reference games with abstract referents remain a difficult task for machine learning models, particularly in the few-shot setting \cite{gulCoGenLearningFeedback2024, chenRetrospectiveLearningInteractions2025}. 
Our results point to the important role of contextual information, but also highlight ways in which current VLMs are limited in bootstrapping their own learning.
Further work is needed to more comprehensively characterise the role of prompt engineering, causal and attentional factors underlying context use, and humans' own sensitivity to context.
Furthermore, our study focused on interpretation, but generating appropriate descriptions is a yet harder task that is worth exploring.
These directions will help us to build language models that are flexible but robust, and that can adapt to a range of tasks and settings as humans do.

\section*{Limitations}

While we attempted to match the task for humans and models as much as possible, some differences still remained; for example, we ignored model continuations which were not one of A--L, whereas human participants were explicitly limited in the range of response options (i.e., they were only able to select one of A--L, and could not offer any other response type). 
Additionally, we tested a limited set of abstract stimuli as a representative example, and focused on English language responses; generalisability to other (e.g., naturalistic) stimuli sets as well as other languages (especially low-resource languages) is left to future work. 
Finally, while iterated reference games serve as a paradigm for important kinds of pragmatic inference, it is not a pure measure of pragmatics, nor is it a comprehensive evaluation for pragmatic ability and reasoning in general; other complementary tasks will be needed to more fully assess the pragmatic capabilities of any model.


\section*{Acknowledgments}

We would like to thank Michael C. Frank for advice and funding support, as well as members of the Language and Cognition Lab, members of the Social Interaction Lab, and reviewers at ARR and the NeurIPS CogInterp Workshop for helpful feedback on this work.

\bibliography{custom}

\newpage
\appendix

\section{Additional methodological details}\label{app:add_methods}

\subsection{Human experiments}\label{app:human}

Human experiments were approved by Stanford University IRB under protocol number 20009. 
Data were collected in August 2025; for reference, data from the original games in \citet{boyceInteractionStructureConstrains2024} were collected in June 2024, and na\"{i}ve human data from \citet{boyceIdiosyncraticNotOpaque2025} were collected in November 2024. 
Participants were recruited via Prolific and compensated \$3.50 for 20 minutes of their time (hourly wage \$10.50, which is above the minimum wage in the United States), as well as \$0.05 for each correct answer as a bonus (up to \$3.60). 
Participants had to be fluent in English, and not have participated in prior related studies. 
Before beginning the experiment, participants gave informed consent about participating in the experiment, that they were assured anonymity, and that they could decline participation at any time.
All responses were anonymised. 


The following instructions were presented to participants at the start of the experiment:

\begin{tcolorbox}
    \small
    \begingroup
    \setlength{\parskip}{4pt}
    \textbf{Please read these instructions carefully!}
    
    In this task, you will see a \textbf{transcript of a conversation} some previous participants had where a \textbf{speaker identified one of the images to some listeners}.

    Your goal is to \textbf{read the transcript} and \textbf{figure out which image} is being described!

    You will \textbf{read the transcript word-by-word, pressing SPACEBAR to reveal each word.} When you have finished reading, you will \textbf{click the image} you think is being described. Then you will \textbf{find out if your selection is correct}.

    You will see \textbf{72 trials}. The image choices will be the same on each trial, but the descriptions may come from different groups of people.

    Some of the transcripts may be much shorter or longer than others. Some of the descriptions may be hard to understand -- just take a guess if you are not sure.

    You will get a bonus of \textbf{5 cents} for each image you get right! 

    \textbf{Note:} These descriptions came from real people and may contain inappropriate or offensive language. If you find a description inappropriate, please let us know via Prolific message or in the exit survey, so we can filter it out in the future.
    \endgroup
\end{tcolorbox}









 

On each trial, participants saw instructions ``Press space to reveal the next word.'' until they had revealed all the text, when the instructions switched to ``Click on the image that was described.'' 
  
\subsection{Prompt and in-context trials}

The system prompt for all language model generations included the following text:

\begin{tcolorbox}
    \small
    You will be presented with a list of messages between people playing a reference game, where the describer has to get the matcher to choose a shape from a set of shapes. Your goal is to guess which of the shapes the describer is trying to get the matcher to choose. The shapes, with their labels, are shown in the image.
Please answer with just the letter corresponding to the image you think the describer is trying to get the matcher to choose, and no other text. You will receive feedback telling you whether your choice was correct or incorrect.
\end{tcolorbox}

\begin{figure}[th]
    \centering
    \includegraphics[width=.8\linewidth]{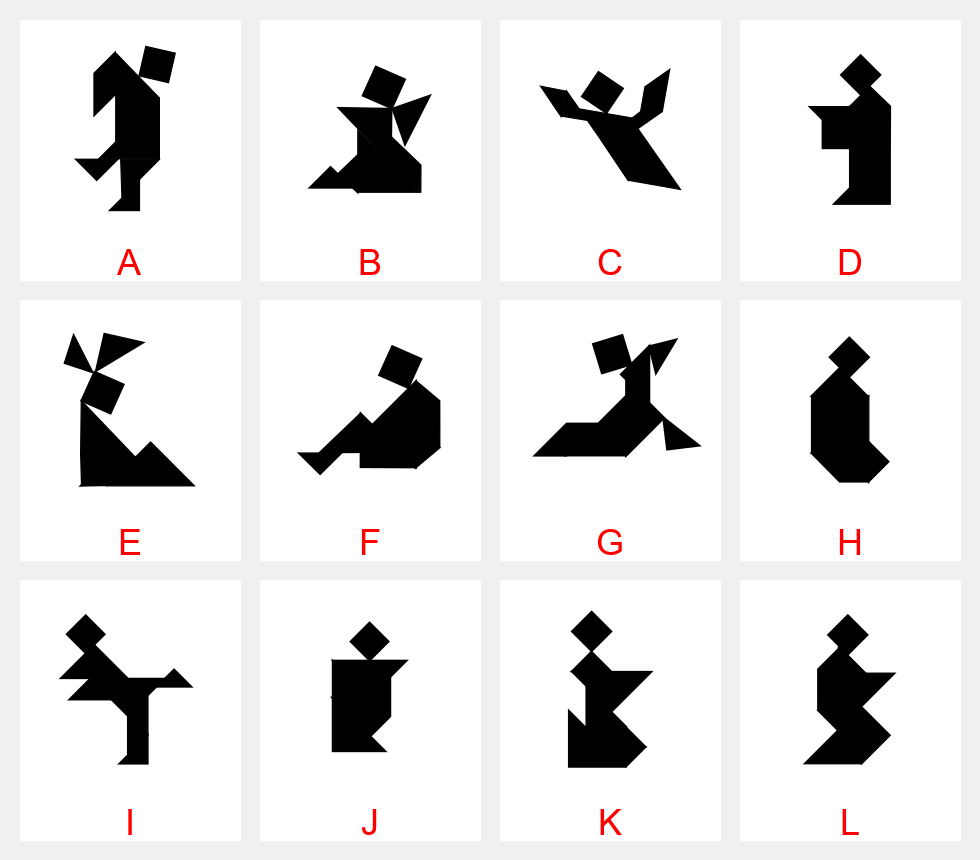}
    \caption{The image that was presented to the vision--language models, containing all 12 tangram shapes with their letter labels.}
    \label{fig:tangrams}
\end{figure}

\begin{figure}
    \centering
    \includegraphics[width=.68\linewidth]{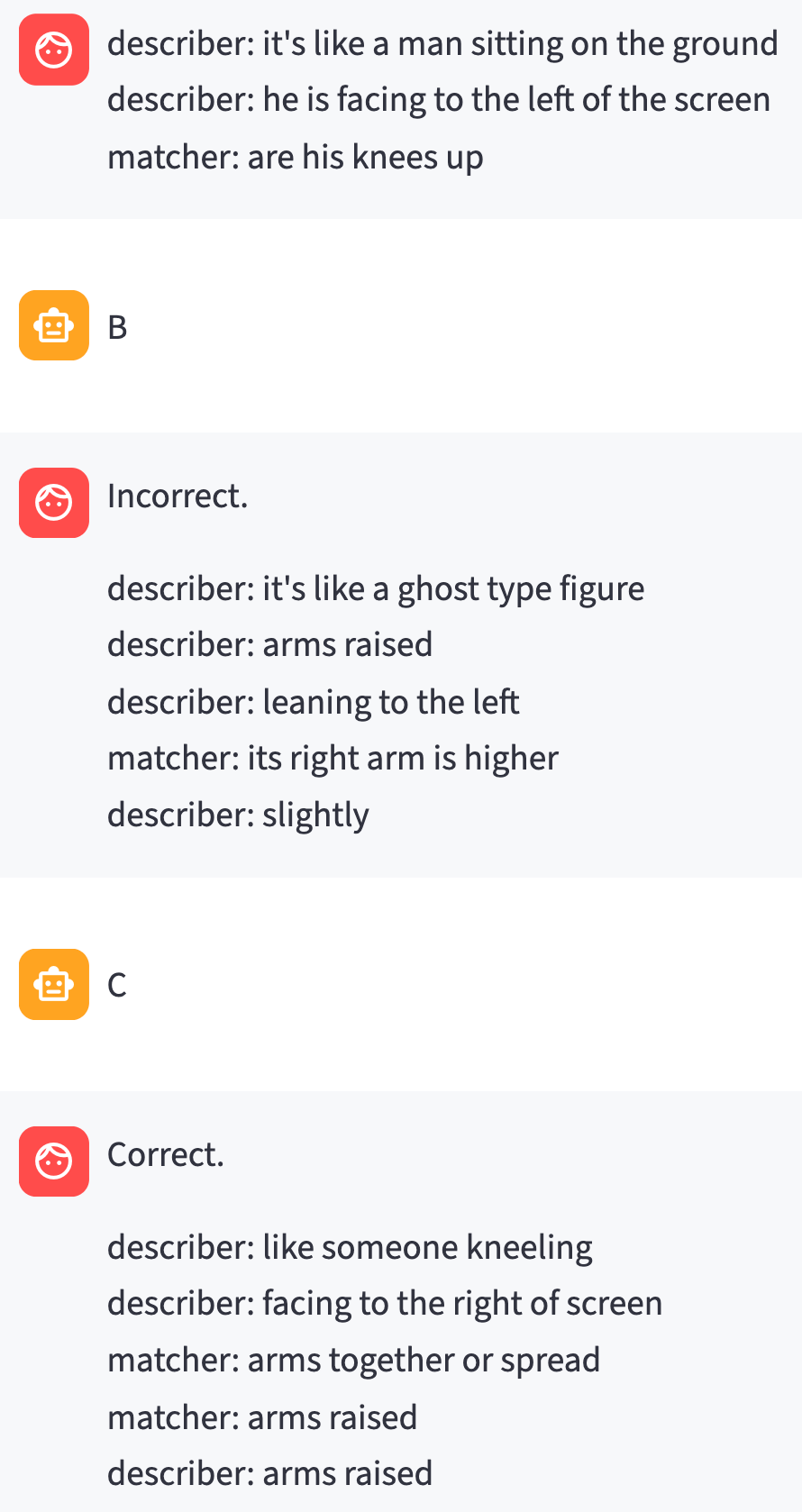}
    \caption{Example of the in-context trials presented to a model, with the \textcolor{red}{User} providing the trial text, the \textcolor{orange}{Assistant} providing a guess, and the \textcolor{red}{User} responding with the correctness of the guess.}
    \label{fig:incontext}
\end{figure}

All conditions except for the no-image conditions included the image in Figure~\ref{fig:tangrams}.

In-context trials were presented as if the Assistant had responded with a single letter guess, and the User then responds with whether the guess was correct or incorrect, as shown in Figure~\ref{fig:incontext}. 

\subsection{Inference details}\label{app:inference}

We ran open-weights models with the vLLM inference framework \citep{kwon2023efficient}. Models were run in bfloat16 precision on two NVIDIA A40 GPUs. We generated the top 1000 logprobs for the first completion token from these models. In some conditions, some of the open-weights models assigned low probability to all uppercase letters. For instance, Molmo2 preferred lowercase letters in the no image no context condition. For consistency, we use the renormalized distribution over capital letters A-L in all cases.

We called frontier models using the corresponding APIs: the OpenAI Responses API, the Anthropic API, and the Google GenAI API for GPT 5.2, Claude 4.6 Sonnet, and Gemini 3 Flash, respectively. For each of these models, we turned the reasoning level to the lowest possible setting. We sampled ten completions of at most 256 tokens, then parsed each completion for an uppercase letter representing the model's choice. If the model did not respond with just a single uppercase letter, the parser first searches for and removes any text that looks like the model continued the conversation, like a model-generated ``Correct.'' or ``describer:''. Next, the parser checks for an answer at the start or the end of the remaining completion.

\subsection{Span annotation}\label{app:annotation}
We prompted Olmo 3 7B Instruct \cite{olmo2025olmo3} to identify spans and tag them along two dimensions: literal / metaphorical, and part-based / holistic.
One author (AWMT) manually corrected the spans and tags.
Another author (VB) independently corrected the spans and tags for 20\% of the trials to determine inter-rater reliability.
For each trial, we calculated the proportion of spans tagged as metaphorical, and the proportion tagged as holistic.
We calculated the single score two-way agreement intraclass correlation coefficient for each dimension \cite{mcgrawFormingInferencesIntraclass1996}.
Inter-rater reliability was excellent for both dimensions, ICC(A, 1) = .954 [.937, .967] for metaphorical, ICC(A, 1) = .824 [.764, .870] for holistic.

\section{Additional results}
\label{app:add_results}

\subsection{Performance varies by original trial number}

\begin{figure}[t!]
    \centering
    \includegraphics[width=\linewidth]{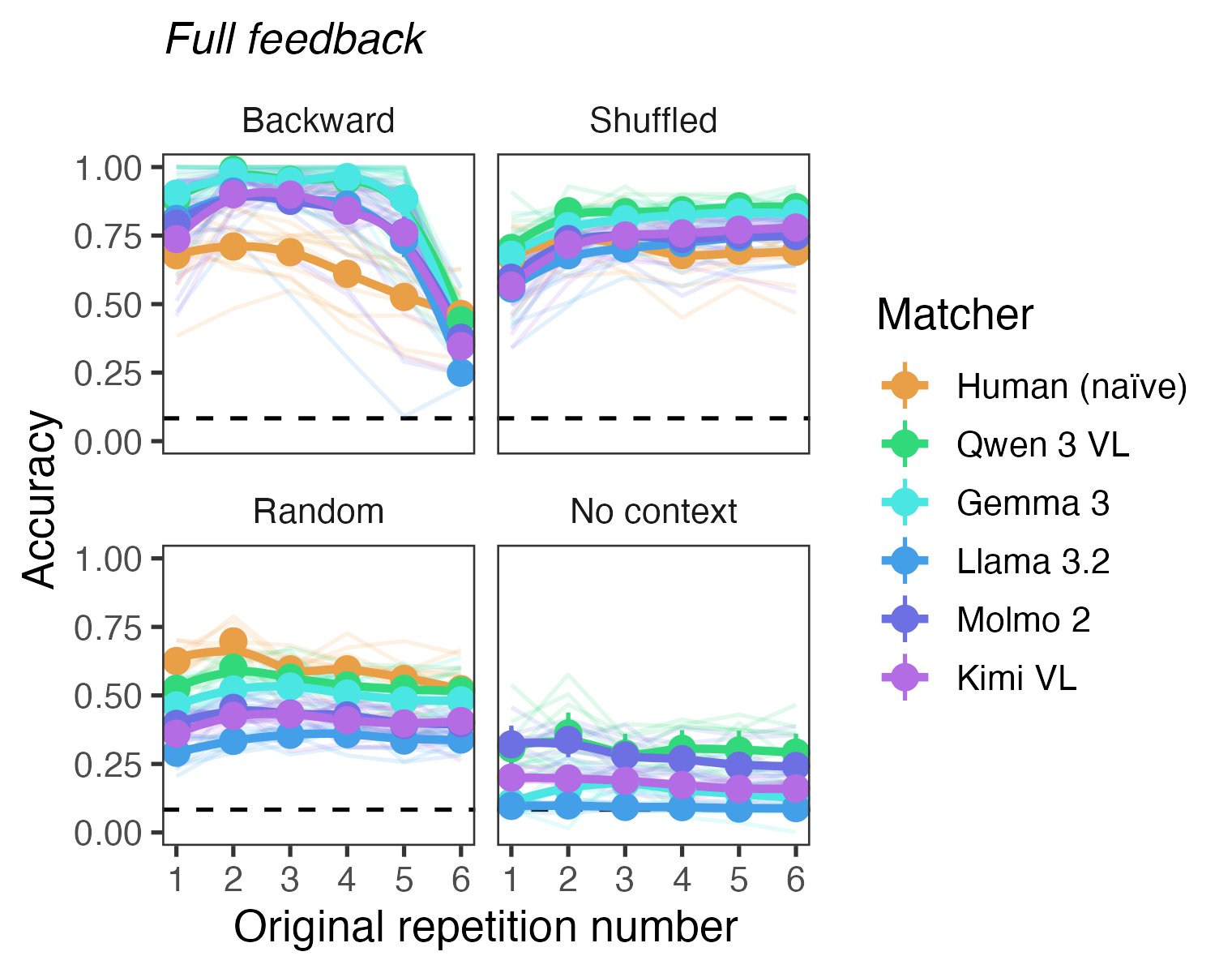}
    \caption{Matcher accuracy for the shuffled, backward, random, and no context conditions across all matcher types (both human and model), with best-fit LOESS curves, shown by repetition number from the original game. Error bars indicate bootstrapped 95\% confidence intervals. Dashed lines indicate the chance level (0.083). }
    \label{fig:original}
\end{figure}

Four conditions displayed trials in a different order than the original games: shuffled, backward, random, and no context.
To understand the effect of the original trial number, we replotted matcher performance for these conditions by original repetition number in Figure~\ref{fig:original}.\footnote{Note that the results for the no context condition are identical to those in Figure~\ref{fig:noncumulative} as they were shown by original repetition number in that plot.}
The general worsening performance over rounds across these conditions suggests that increasing conventionalisation in human referring expressions leads to more idiosyncratic expressions \citep[as in][]{boyceIdiosyncraticNotOpaque2025} that are harder to resolve.
Interestingly, humans (and some models) showed highest performance in the second repetition (rather than the first); this trend was also observed in the ablated condition results in Figure~\ref{fig:noncumulative}.
This phenomenon suggests that there might be a trace of practice effects in human describers, whereby they exhibit increasing systematisation from the first to the second repetition, before showing conventionalisation of expressions.

\subsection{Humans are more correlated with each other than with models\label{app:corrs}}

\begin{table*}[th]
    \small
    \centering
    \caption{Human split-half correlations and model--human correlations for the yoked, shuffled, backward, and random conditions. Human split-half correlations show the mean and 95\% confidence interval over 1000 random splits.}
    \label{tab:corrs}
    \begin{tabular}{lcccccc}
    \toprule
    & \multicolumn{6}{c}{$r$} \\ \cmidrule{2-7}
    Condition & Human split-half   & Qwen 3 VL & Gemma 3 & Llama 3.2 & Molmo 2 & Kimi VL \\ \midrule
    Yoked & $.42\;[.32, .50]$ & $.08$ & $-.07$ & $-.02$ & $.46$ & $.19$ \\
    Backward & $.48\;[.40, .55]$ & $.20$ & $-.07$ & $.02$ & $.40$ & $.19$ \\
    Shuffled & -- & $.23$ & $.02$ & $.16$ & $.47$ & $.31$ \\
    Random & -- & $.23$ & $.04$ & $.11$ & $.59$ & $.29$ \\ \bottomrule
    \end{tabular}
\end{table*}

Correlation values between models and na\"ive humans, as well as human split-half correlations, are shown in Table~\ref{tab:corrs}.
For the shuffled and random conditions, we averaged accuracies for each original trial prior to calculating correlations.
Human split-half correlations were computable only for yoked and backward conditions, in which multiple participants saw the same trials in the exact same order.
We also note that the human split-half correlations are likely underestimated as some splits did not have the whole set of original trials in each split, and thus had some dropped trials.
Overall, models were less similar to humans than humans were to other humans in the yoked and backward conditions; we hypothesise that the same would be true for the shuffled and random conditions if appropriate data were available.

\begin{figure}[t]
    \centering
    \includegraphics[width=\linewidth]{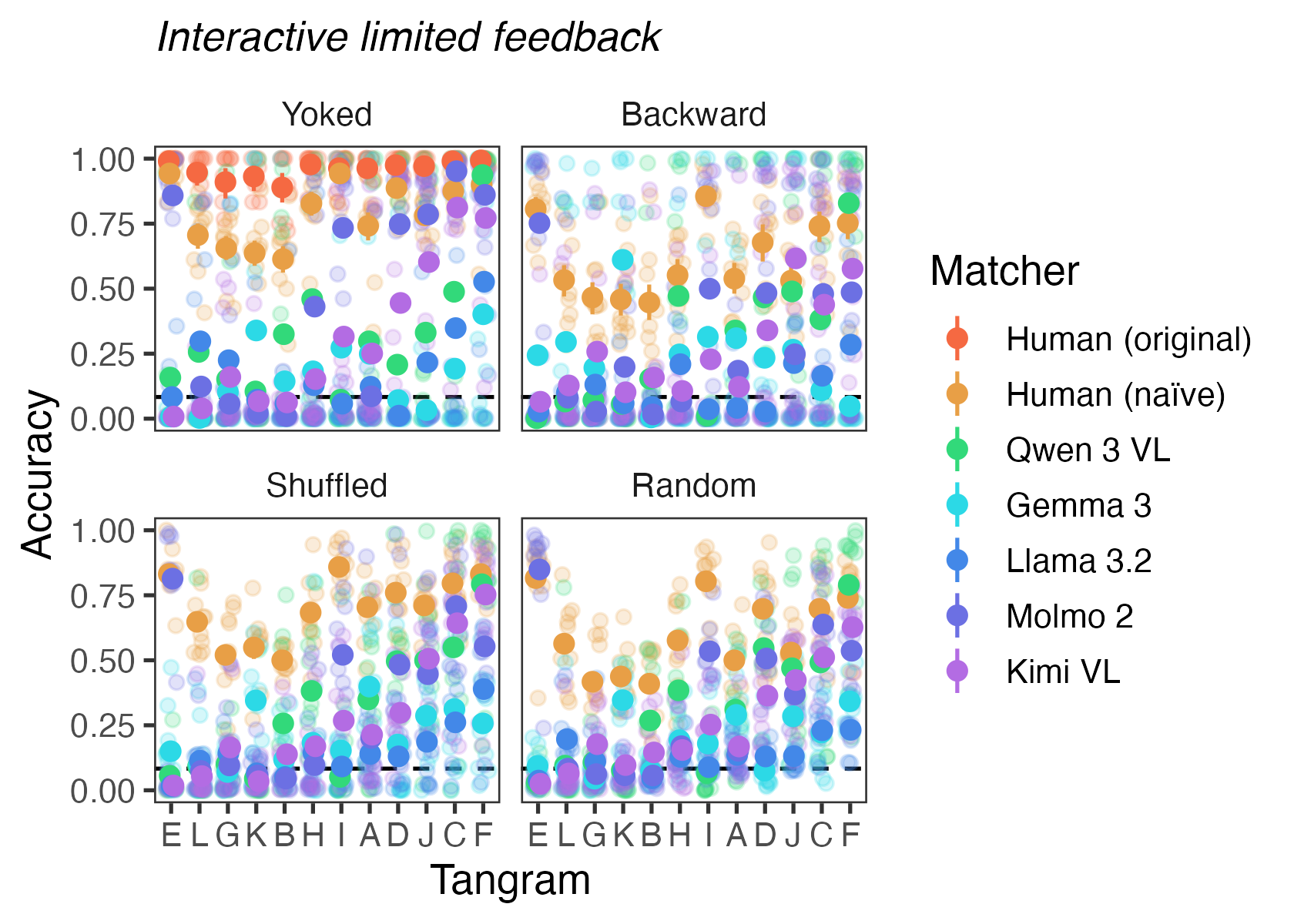}
    \caption{Item-wise variation in accuracy across matchers and conditions. Error bars indicate bootstrapped 95\% confidence intervals. Dashed lines indicate the chance level (0.083).}
    \label{fig:items}
\end{figure}

\subsection{Items show large variation in performance}

Figure~\ref{fig:items} shows the item-wise performance for all matchers and conditions. 
Both naïve humans and models display large item-wise variation, with the magnitude of this variation sometimes exceeding that due to repetition number, matcher, or condition, replicating the item-wise variation found by~\citet{boyceIdiosyncraticNotOpaque2025}.
There appears to be some shared variance across models, although each model also has particular idiosyncrasies and biases (e.g., Molmo 2 appears to be especially accurate for tangram E compared to other models).

\subsection{Conditions pattern with each other systematically}

\begin{figure*}
    \centering
    \includegraphics[width=\linewidth]{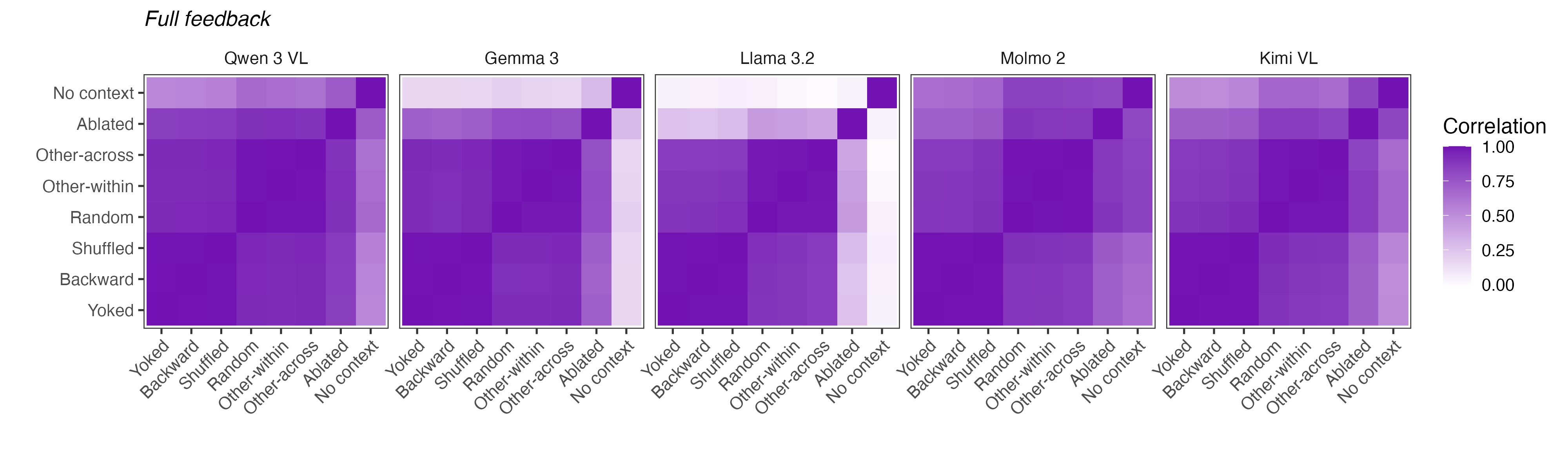}
    \caption{Correlation in confusion matrices across conditions for all models.}
    \label{fig:rsa}
\end{figure*}

Drawing inspiration from representational similarity analysis \cite{kriegeskorteRepresentationalSimilarityAnalysis2008}, we conducted a correlation analysis over the confusion matrices for each model across conditions under full feedback.
We constructed confusion matrices for each model--condition combination by taking the mean of the probability distributions across trials for each target tangram.
We then calculated the Pearson's correlation among the confusion matrices for all conditions.
The resultant correlograms are shown in Figure~\ref{fig:rsa}.
These correlograms demonstrate relatively high correlations across conditions for all models (all $r > .85$), except for the ablated condition ($r$ = .25 -- .91) and the no context condition ($r$ = -.01 -- .85).
Furthermore, all models show a similar pattern of correlations, with the yoked, shuffled, and backward conditions clustering together, the other-within, other-across, and random conditions clustering together, and the ablated and no context conditions being the most distinct from the other conditions.
This pattern of results partly reflects the differences in overall performance across the different conditions, but also supports the idea that relevance is a crucial organising dimension of the content of contextual information.

\begin{figure}
    \centering
    \includegraphics[width=\linewidth]{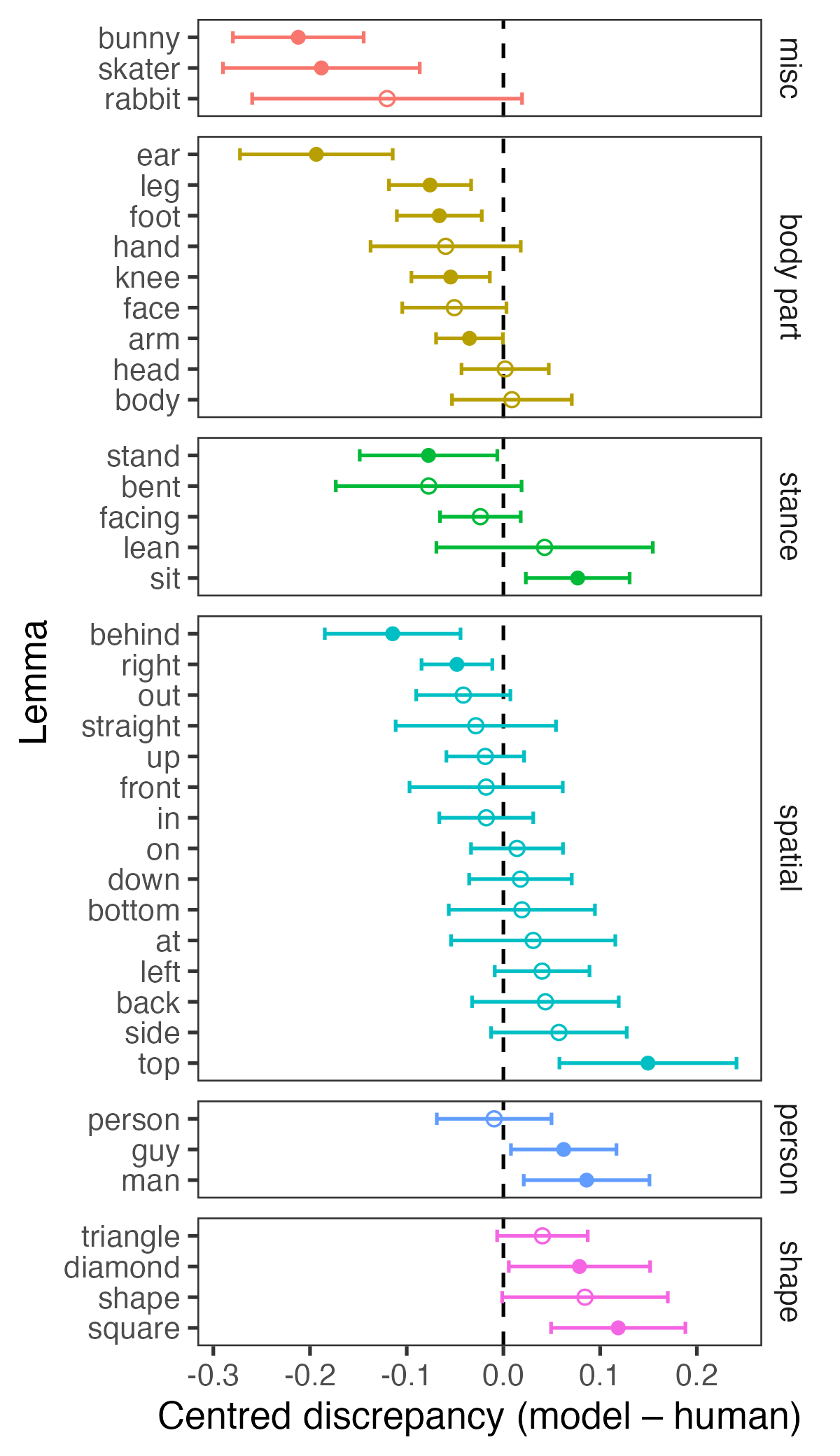}
    \caption{Centred discrepancy in model versus human accuracies in the random condition for selected lemmas, categorised by semantic categories. Error bars indicate 95\% confidence intervals. Dashed line indicates zero (i.e., discrepancy equal to the grand mean). Empty/filled points indicate whether the confidence intervals do/do not include zero respectively.}
    \label{fig:lemmas}
\end{figure}

\subsection{Lemmas are related to model--human discrepancies}\label{app:lemmas}

To understand the discrepancies between models and humans, we used model and na\"ive human responses in the random condition under interactive limited feedback, in which there was limited relevant context, although we note that our method could be applied to other conditions as well.
We estimated the discrepancy between model and human performance by calculating the difference between model and human accuracy for each unique trial, averaging across all models and runs.
Then, we lemmatised the message texts in each trial using UDPipe \cite{strakaUDPipe20Prototype2018}, and estimated the means and 95\% confidence intervals of the model--human discrepancy for each unique lemma across all trials in which that lemma occurred.
We recentred these discrepancy scores by subtracting the grand mean in discrepancy across all trials (-0.36), and determined the significance of the centred discrepancy (CD) estimates by calculating whether the 95\% confidence intervals contained zero. 
These CD values reflected whether models were better or worse than average, in comparison to humans.
We also filtered down to lemmas that occurred in at least 4 games and in at least 20 trials (i.e., their discrepancies were not due to a small number of idiosyncratic games).
The CD values for selected lemmas are shown in Figure~\ref{fig:lemmas}.

Manual inspection of the resulting estimates reflects the role of metaphors in model--human discrepancy. 
Lemmas relating to literal features of the tangram such as geometric shapes (e.g., ``triangle'', ``diamond'', ``square'') and spatial locations (e.g., ``left'', ``side'', ``top'') had CD values that were positive or not different than zero. 
Conversely, words which required more metaphorical interpretations, including ``skater'' and ``bunny'', as well as interpretations of tangram parts as body parts (e.g., ``ear'', ``leg'', ``foot''), had negative CD values.
Interestingly, some metaphorical words including ``guy'' and ``man'' had positive CD values; it is possible that these words were generic enough that they were highly frequent and could be found in many combinations with other lemmas, including with other literal descriptors. 
These results also converge with our span-based analysis of model and human performance in Section~\ref{sec:spans}.

\subsection{Given additional practice, models still fail to improve}\label{app:practice}

The change in models' performance over repetitions is a function of both the change in the human descriptions of the tangrams, as well as models getting better at the task of interpreting descriptions.
To isolate the effect of practice over trials, we tested models on additional conditions in which we repeated the descriptions from a single round of the reference game; thus, the descriptions were held constant, and models had multiple chances to learn from a specific set of descriptions.
We tested two conditions drawing from descriptions that were originally in the first round and the last round, under interactive limited feedback.

\begin{figure}[t]
    \centering
    \includegraphics[width=\linewidth]{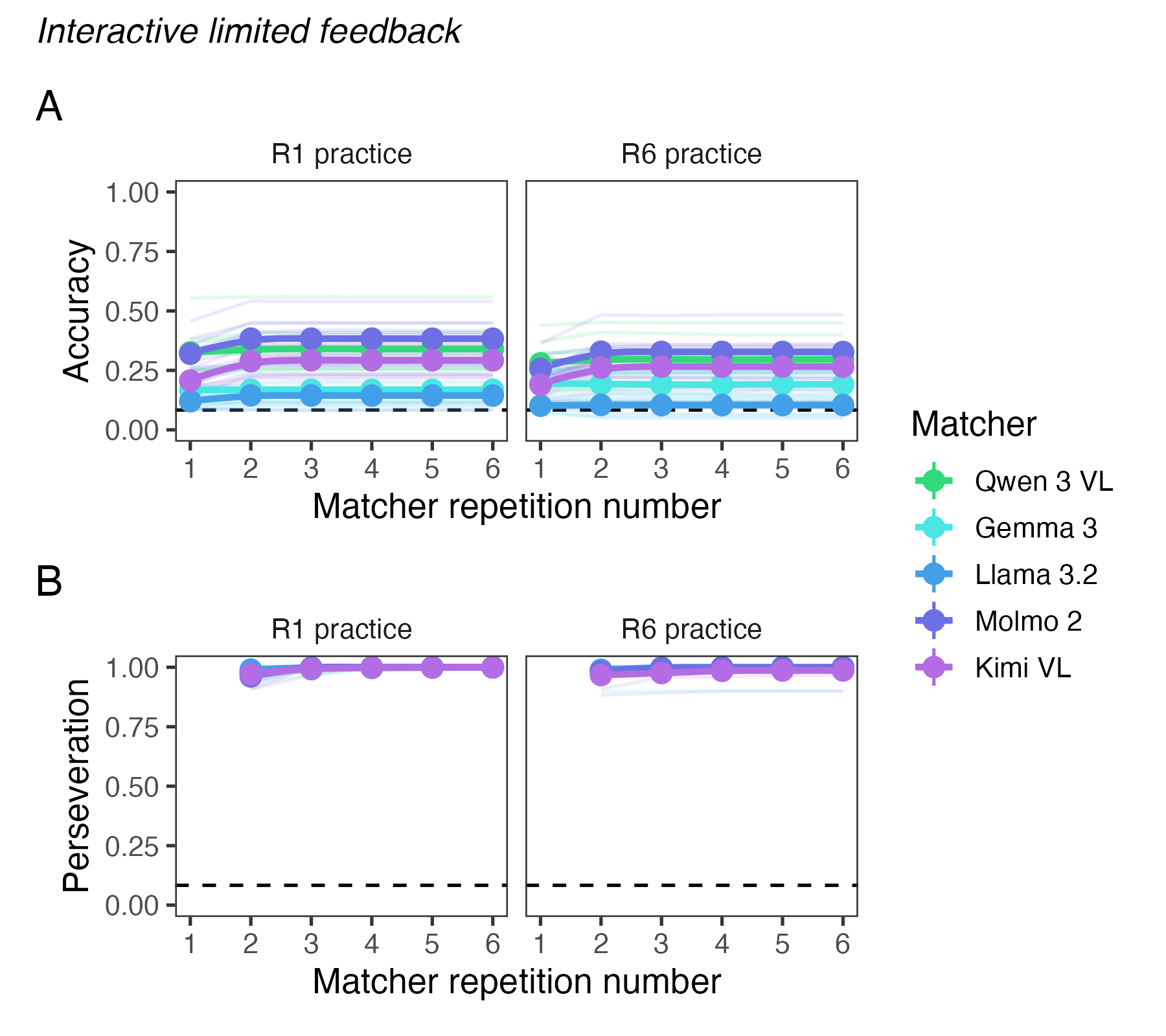}
    \caption{(A) Matcher accuracies and (B) likelihood of perseverating on their previous answer for practice conditions. Error bars indicate bootstrapped 95\% confidence intervals. Dashed lines indicate the chance level (0.083).}
    \label{fig:practice}
\end{figure}

As shown in Figure~\ref{fig:practice}A, models performed extremely poorly in the practice conditions, revealing that they do not learn adequately with practice.
After a small initial boost from the first to the second rounds, performance plateaus at <0.4. 
We then checked whether models' poor performance was due to perseverating on their own previous incorrect answer; to do so, we examined whether the response with the highest log probability also had the highest log probability in the previous repetition (including ties). 
Figure~\ref{fig:practice}B shows that this trend was indeed the case, with all models perseverating almost 100\% of the time. 
In fact, perseveration went up slightly over repetitions, suggesting that instead of learning, models became more entrenched in incorrect responses.
Together with results from Section~\ref{sec:learning}, these patterns show that invalid in-context information significantly worsens model performance.

\begin{figure}[t]
    \centering
    \includegraphics[width=\linewidth]{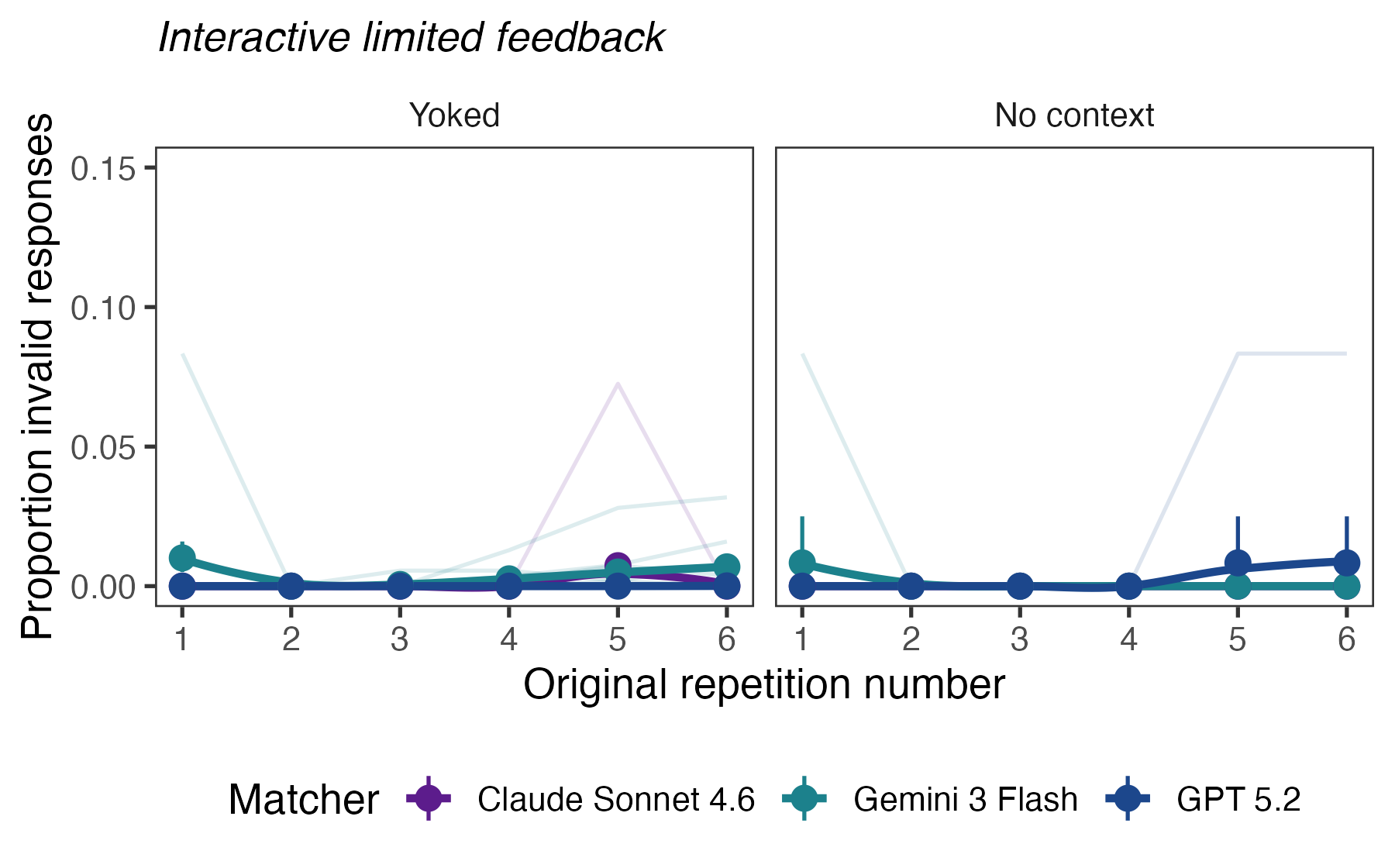}
    \caption{Proportion of invalid responses for frontier models under interactive limited feedback.}
    \label{fig:invalid}
\end{figure}

\subsection{Frontier models sometimes, but rarely, return invalid responses}\label{sec:invalid}
As mentioned in \ref{app:inference}, for frontier models, we parsed the responses to ignore non-answer text after the model's response.
However, on occasion, models would fail to return a parsable answer in 256 tokens, leading to invalid responses.
These were ignored when calculating response probabilities and accuracies.
Figure~\ref{fig:invalid} shows the rate of invalid responses across our frontier models, which were relatively low overall. 

\end{document}